\documentclass{article}

\usepackage{microtype}
\usepackage{graphicx}
\usepackage{subfigure}
\usepackage{booktabs} 

\usepackage{hyperref}

\usepackage[usenames,dvipsnames]{xcolor}
\definecolor{commentcolor}{RGB}{110,154,155}   

\usepackage{caption,dcolumn,adjustbox, multirow, array}  
\newcolumntype{d}[1]{D{.}{.}{4}}
\usepackage{caption}

\usepackage{graphicx}
\graphicspath{{./figs/}}

\usepackage[accepted]{icml2025}

\usepackage{amsmath}
\usepackage{amssymb}
\usepackage{mathtools}
\usepackage{amsthm}
\usepackage{microtype}      
\hypersetup{
    colorlinks,
    linkcolor={red!50!black},
    citecolor={blue!50!black},
    urlcolor={blue!80!black}
}
\usepackage[capitalize,noabbrev]{cleveref}

\theoremstyle{plain}

\theoremstyle{definition}

\theoremstyle{remark}

\usepackage[textsize=tiny]{todonotes}

\icmltitlerunning{Wearable Accelerometer Foundation Models for Health via Knowledge Distillation}

\begin{document}

\twocolumn[
\icmltitle{Wearable Accelerometer Foundation Models for Health via Knowledge Distillation}



\icmlsetsymbol{equal}{*}

\begin{icmlauthorlist}
\icmlauthor{Salar Abbaspourazad}{apple}
\icmlauthor{Anshuman Mishra}{apple}
\icmlauthor{Joseph Futoma}{apple}
\icmlauthor{Andrew C Miller}{apple}
\icmlauthor{Ian Shapiro}{apple}
\end{icmlauthorlist}

\icmlaffiliation{apple}{Apple Inc}

\icmlcorrespondingauthor{Salar Abbaspourazad}{salarabb@apple.com}
\icmlcorrespondingauthor{Ian Shapiro}{ishapiro@apple.com}

\icmlkeywords{Machine Learning, ICML}

\vskip 0.3in
]



\printAffiliationsAndNotice{}  

\begin{abstract}
Modern wearable devices can conveniently record various biosignals in the many different environments of daily living, enabling a rich view of individual health. However, not all biosignals are the same: high-fidelity biosignals, such as photoplethysmogram (PPG), contain more physiological information, but require optical sensors with a high power footprint. Alternatively, a lower-fidelity biosignal such as accelerometry has a significantly smaller power footprint and is available in almost any wearable device. While accelerometry is widely used for activity recognition and fitness, it is less explored for health biomarkers and diagnosis. Here, we show that an accelerometry foundation model can predict a wide variety of health targets. To achieve improved performance, we distill representational knowledge from PPG encoders to accelerometery encoders using 20 million minutes of unlabeled data, collected from $\sim$172K participants in the Apple Heart and Movement Study under informed consent. We observe strong cross-modal alignment on unseen data, e.g., $99.2\%$ top-1 accuracy for retrieving PPG embeddings from accelerometry embeddings. We show that distilled accelerometry encoders have significantly more informative representations compared to self-supervised or supervised encoders trained directly on accelerometry data, observed by at least 23\%-49\% improved performance for predicting heart rate and heart rate variability. We also show that distilled accelerometry encoders are readily predictive of a wide array of downstream health targets, i.e., they are generalist foundation models. We believe accelerometry foundation models for health may unlock new opportunities for developing digital biomarkers from any wearable device.
\end{abstract}

\section{Introduction}

The recent growth of wearable devices has empowered individuals to track their health more conveniently and frequently. Wearable devices can record various biosignals and provide individuals with a continuous view of their health metrics in daily environments, a view which is hard to achieve in clinical settings. While this unlocks new exciting opportunities for wearable devices, challenges remain. High-fidelity biosignals, such as photoplethysmogram (PPG) and electrocardiogram (ECG), have substantial physiological information content and are commonly used for developing digital biomarkers. However, they require specific hardware and sensors for recording, therefore are unavailable in some wearable devices. Even when available, it is difficult to collect them frequently due to power constraints of small wearable devices, or reliance on active user engagement for recording (e.g., ECG). On the other hand, a lower-fidelity biosignal, such as accelerometry, is available on
almost all wearable devices and has a significantly small power footprint, making it ideal for
enabling digital biomarkers efficiently, frequently, and on a variety of wearable devices. 

Accelerometry during gross motion is widely used for activity recognition and fitness, where it captures active movements of the body and not direct physiological information. In turn, during low-motion periods, accelerometry is capable of detecting minute physiological movements on the body as a result of heart function, known as ballistocardiogram  \citep{inan_ballistocardiography_2015, kim_ballistocardiogram_2016}, therefore, it contains complex physiological information. Despite this, accelerometry is less explored for health diagnosis and digital biomarkers on wearable devices compared to other biosignals such as PPG, given that its relationship to diverse health targets is not as well-understood and that it is more susceptible to external artifacts. Examples of how less-structured and noisier accelerometry is compared to PPG can be found in Figure \ref{fig: methods}. Here, we address this research question by training a generalist foundation model for accelerometry during low-motion, in a large-scale dataset collected under natural living conditions, and demonstrate that accelerometry can predict a diverse set of health targets. See Section \ref{section: related_work} for how our work relates to the existing work. 

Using unlabeled sensor data collected under informed consent from the large longitudinal Apple Heart and Movement Study (AHMS) \citep{macrae_apple_2021, truslow_understanding_2024}, we train generalist accelerometry foundation models via cross-modal representational knowledge distillation from PPG. Our contributions are: 1) \textbf{Accelerometry foundation models for health}: We train accelerometry foundation models for health via representational knowledge distillation from PPG using a large-scale dataset with 20M minutes of multi-modal sensor data from $\sim$172K participants. For the first time, we show that a single accelerometry encoder is predictive of a wide array of health targets with a high accuracy. 2) \textbf{Fully unsupervised representational knowledge distillation framework}: While multi-modal modeling and cross-modal reconstruction of biosignals have been done before, we combine and adopt techniques inspired by uni-modal and multi-modal pre-training frameworks from other domains of deep learning to create a fully unsupervised distillation framework for biosignal time-series. Knowledge distillation has been used for improved information extraction from accelerometry, and our end-to-end method being unsupervised is crucial for health applications where labeled data is often limited. 3) \textbf{Studying representational alignment}: We study representational alignment of PPG and accelerometry by doing retrieval analysis. 4) \textbf{Studying representational information}: We study the representational power of accelerometry encoders for critical health targets such as heart rate and heart rate variability, across different available labeled data regimes, as well as for demographic variables and 46 health targets including health conditions, use of medications and lifestyle habits from AHMS survey questions. In addition to these contributions, to support the robustness of the findings, we perform ablation studies to evaluate the efficacy of the encoder architecture (Transformer and EfficientNet), the pre-training strategy of the PPG teacher model (masked autoencoding and contrastive learning), the augmentations, and other training choices.

\section{Related work}
\label{section: related_work}
\textbf{Uni-modal and multi-modal representation learning}: Unsupervised representation learning techniques, e.g., self-supervised learning, have been proven successful in training generalist models, also known as foundation models, without requiring any explicit labels during training in various domains of deep learning such as natural language processing \citep{devlin_bert_2019, openai_gpt-4_2023}, computer vision \citep{chen_simple_2020, chen_empirical_2021, he_masked_2021, oquab_dinov2_2023}, speech recognition \citep{baevski_wav2vec_2020, baevski_data2vec_2022}, and health \citep{cheng_subject-aware_2020, kostas_bendr_2021, sarkar_self-supervised_2022, mohsenvand_contrastive_2020, gopal_3kg_2021, kiyasseh_clocs_2021,mehari_self-supervised_2022, wu_representation_2020, spathis_self-supervised_2021, tang_exploring_2021, yuan_self-supervised_2023, lai_practical_2023,abbaspourazad_large-scale_2024,liu_frequency-aware_2024, narayanswamy_scaling_2024}. While most of these works have been primarily on training uni-modal foundation models, there has been a recent shift in training multi-modal foundation models to allow for leveraging information from multiple modalities, either to train a model that simultaneously processes multiple modalities \citep{mizrahi_4m_2023,meta_chameleon_2024}, or to train and bind multiple modality-specific foundation models \citep{radford_learning_2021, girdhar_imagebind_2023, thapa_sleepfm_2024}, particularly with a contrastive objective. Similarly for health applications, there has been a growing interest in cross-modal reconstruction of biosignals \citep{sarkar_cardiogan_2020}, or simultaneously pre-training multiple biosignal modalities \citep{deldari_cocoa_2022, liu_frequency-aware_2024, deldari_crossl_2024, thapa_sleepfm_2024} or their extracted features \citep{narayanswamy_scaling_2024}.

\textbf{Knowledge distillation}: Knowledge distillation has been extensively used for transferring knowledge from a neural network (teacher) to another neural network (student) in other domains \citep{hinton_distilling_2015, tian_contrastive_2022}, traditionally often in supervised settings to transfer knowledge from a large neural network to a small neural network \citep{hinton_distilling_2015, tian_contrastive_2022} or from a high-fidelity modality to a low-fidelity modality \citep{gupta_cross_2015, aytar_soundnet_2016, tian_contrastive_2022}, or from an ensemble network into a single one \citep{tian_contrastive_2022}, by using the teacher's output logits as soft labels of the student model. Alternatively, knowledge distillation can also be performed using intermediate representations. With the recent emergence of foundation models, there has been several works for combining self-supervised learning and knowledge distillation via self-distillation \citep{xie_self-training_2020, fang_seed_2021, caron_emerging_2021}, or distilling one or several existing foundation models to a single foundation model to improve and agglomerate their representations \citep{ wei_contrastive_2022, wang_sam-clip_2024, ranzinger_am-radio_2024}.

In this work, we adopt ideas from different related work in uni-modal and multi-modal representation learning and knowledge distillation, to distill knowledge from a foundation model for a high-fidelity modality in order to get a foundation model for a low-fidelity modality. Similar to uni-modal foundation models, we train our PPG teacher encoder using masked autoencoding \citep{he_masked_2021} and contrastive learning \citep{chen_simple_2020}, with our own variation of these frameworks. We then transfer knowledge from our PPG teacher encoders to accelerometry student encoders via a cross-modal knowledge distillation framework \citep{tian_contrastive_2022} that bears similarity to multi-modal contrastive learning \citep{radford_learning_2021, girdhar_imagebind_2023} and self-distillation for model compression \citep{fang_seed_2021}. There has been prior work on training uni-modal foundation models for PPG and accelerometry \citep{spathis_self-supervised_2021, yuan_self-supervised_2023, abbaspourazad_large-scale_2024, pillai_papagei_2024} or multi-modal foundation models for other biosignals \citep{deldari_cocoa_2022, liu_frequency-aware_2024, deldari_crossl_2024, thapa_sleepfm_2024}. These multi-modal biosignal foundation models either take multiple modalities as input \citep{liu_frequency-aware_2024, deldari_crossl_2024}; or they ``bind'' multi-modality embeddings into a shared subspace \citep{deldari_cocoa_2022,thapa_sleepfm_2024} similar to \citep{radford_learning_2021} from randomly-initialized encoders, which is primarily different for our motivation of cross-modal representational knowledge distillation and model compression from an existing high-fidelity teacher encoder. In fact, we show that training both encoders of high-fidelity and low-fidelity modalities together degrades the quality of low-fidelity embeddings (see Section \ref{section: extra_ablations}). In line with our motivation, there has been prior work on leveraging asymmetric information in biosignals, by cross-modal reconstruction of ECG from PPG, for a more accurate estimation of heart rate \citep{sarkar_cardiogan_2020}. We emphasize that our work focuses on accelerometry during low-motion periods where it captures minute cardiovascular signals such as the ballistocardiogram, which is significantly distinct from modeling accelerometry during gross motion for activity recognition, health and fitness \citep{hallgrimsson_learning_2018, ni_modeling_2019, spathis_self-supervised_2021, xu_relcon_2024}. Also, self-supervised learning has recently been shown useful for predicting sleep stages from accelerometry \citep{yuan_self-supervised_2023}. All in all, we develop a foundation model for accelerometry during low motion, by distilling embeddings from a pre-trained PPG teacher foundation model, and demonstrate that it is readily predictive of a wide array of downstream health targets.

\section{Methods and implementation details}

Our knowledge distillation framework is fully unsupervised and consists of two steps: teacher pre-training and cross-modal representational knowledge distillation as depicted in Figure \ref{fig: methods}. Below, we go over the details of each of these steps as well as their implementation details.

\subsection{Teacher pre-training}
\label{section: teacher_pre_training}
The first step of our representational knowledge distillation framework is to pre-train the PPG teacher encoder via self-supervised learning. To show that our knowledge distillation framework is agnostic to the pre-training of the PPG teacher encoder and study its efficacy, we investigated two popular pre-training strategies: masked autoencoding (MAE) as the main framework and contrastive learning (CL) as an ablation, which we explain in detail below.

\textbf{Masked autoencoding}: Our masked autoencoding pre-training framework is adopted from the prior work on images \citep{he_masked_2021}, for time-series. We turn the multi-channel PPG input (4-channels and 60s-long, as explained in Section \ref{section: datasets}) into patches using non-overlapping fixed-length windows, and then project the patches with a learnable linear tokenizer into tokens, which results in 192 256-D tokens. Sinusoidal positional embeddings are added to the input tokens, then 80\% of input tokens are randomly dropped and the 38 kept tokens are passed through the encoder Transformer to get encoder output tokens. Learnable mask tokens, initialized with a 256-D token drawn from a uniform distribution $\sim U(0,1)$, are added back to the encoder output tokens at positions where input tokens where dropped, followed by adding sinusoidal positional embeddings. These 192 new tokens are then processed by the decoder Transformer to generate the decoder output tokens. Finally, with a linear projection, the decoder output tokens are projected to the multi-channel PPG output, with the objective of reconstructing PPG ``pixel'' values in those indices whose patches/tokens were dropped. For maximum learning rate, we used 2e-4 and batch size was set to 512. A complete list of other architectural and training hyperparameters that are shared across methods using Transformers can be found in Appendix Table \ref{table: transformer_hps}. As a baseline for our distilled accelerometry encoders, we also train masked autoencoders for accelerometry in the same way explained above.

\textbf{Contrastive learning}: While our main pre-training framework for the teacher encoder is masked autoencoding, we perform ablation in regards to the teacher pre-training method and architecture via contrastive learning with EfficientNets. We emphasize that contrastive learning can also be done with Transformer models without any meaningful difference  (Appendix Table \ref{table: ablation_vit_cl}), but we choose EfficientNets for the main results to simultaneously demonstrate an ablation on changing the teacher pre-training strategy and student/teacher architecture. Our contrastive pre-training framework closely follows a prior work for PPG signals \citep{abbaspourazad_large-scale_2024}. We select positive pairs as two augmented views of the same sample to enforce the encoders to contain more segment-level information necessary for the main downstream targets used in this study (Appendix Table \ref{table: ablation_positive_pairs}).
During pre-training, we augment each sample twice with our stochastic augmentation module that consists of a stochastic cascade of several standard individual augmentations \citep{iwana_empirical_2021} including \{cut out: 0.4, magnitude warp: 0.25, add Gaussian noise: 0.25, channel permute: 0.25, time warp: 0.15\}, where the values are the assigned probabilities of whether the augmentation is applied for each segment or not. The two augmented views are then passed through a joint-embedding architecture, where one encoder is an exponential moving average ($m=0.99$ for momentum updates) of the other encoder that is updated via backpropagation. The 256-D embeddings of each encoder are then projected to a lower dimensional subspace (128-D) via multi-layer perceptron projection heads. The training objective is maximizing the mutual information of the down-projected embeddings of the two views of the same PPG segment, and minimizing that for different PPG segments, implemented by a regularized InfoNCE loss. We use Kozachenko-Leonenko (KoLeo) differential entropy estimator \citep{sablayrolles_spreading_2019, oquab_dinov2_2023} with the weight of $0.1$ for regularization, and temperature of $0.04$ for scaling similarity scores in the InfoNCE loss. For the maximum learning rate, we used 1e-3, while batch size was set to 256. Other common hyperparameters for training EfficientNets is available in Appendix Table \ref{table: effnet_hps}. As a baseline for our distilled accelerometry encoders, we also train contrastive learned encoders for accelerometry in the same way explained above.

\subsection{Cross-modal representational knowledge distillation}
\label{subsection:methods_kd}
After pre-training the PPG encoder in Section \ref{section: teacher_pre_training}, we distill its embeddings to an accelerometry encoder in a dataset of paired PPG-accelerometry segments (Section \ref{section: datasets}). This second stage is also fully unsupervised without requiring any explicit labels. We perform this representational knowledge distillation using multi-modal contrastive learning similar to a technique used previously to supervise an image encoder with text (CLIP) \citep{radford_learning_2021}, but here we use it to transfer knowledge. To do this, unlike standard approaches \citep{radford_learning_2021}, we first perform augmentations on both modalities, PPG and accelerometry, using our stochastic augmentation (see Section \ref{section: teacher_pre_training}), where the augmentations are independently drawn for each modality in a PPG-accelerometry pair. We found that augmentations were crucial for the quality of the embeddings (see Ablation \ref{section: extra_ablations}). The augmented PPG and accelerometry signals are then processed by the PPG teacher encoder (frozen) and accelerometry student encoder, respectively, to get 256-D output embeddings. To calculate the objective, we first down-project the embeddings to 128-D with trainable multi-layer perceptron projection heads (one 1024-D hidden layer) for both the student and teacher encoders. We found separate learnable projection heads to a smaller subspace was necessary to avoid representation collapse. The student encoder is trained to generate embeddings similar to the teacher encoder, where the objective is contrastive and maximizes the mutual information of paired PPG and accelerometry embeddings, while minimizing the mutual information of an accelerometry embedding with other PPG embeddings. For each batch of embeddings $h$ from N positive pairs for student and teacher ($h_t$, $h_s$), we define multi-modal InfoNCE, where the teacher embeddings are selected as anchors: $L^{(t \rightarrow s)}_{\texttt{contrastive}} = -\frac{1}{N} \sum_{i=1}^{N} \log \frac{\exp(sim(h_{t}^{i}, h_{s}^{i}) / \tau)}{\sum_{j=1}^{N} \exp(sim(h_{t}^{i}, h_{s}^{j}) / \tau)}
$. Here, $sim(\cdot, \cdot)$ is the cosine similarity function and $L^{(t \rightarrow s)}_{\texttt{contrastive}}$ can be viewed as a $N$-way classification problem ($N =$ batch size), such that $h_s^i$ is the correct pair to $h_t^i$ compared to all other potential pairs in the batch $\{h_s^j | 1\leq j \leq N, j \neq i\}$. The final objective is computed as the weighted sum of InfoNCE, from teacher to student and from student to teacher:

\begin{equation}
\label{eq: objective}
L = \lambda L^{(t \rightarrow s)}_{\texttt{contrastive}} + (1-\lambda)L^{(s \rightarrow t)}_{\texttt{contrastive}},
\end{equation}

where $\lambda$ is the scalar weight between 0 and 1. In our experiments, the temperature parameter is set to 0.04, and unless otherwise specified we set $\lambda$ to 1, to emphasize more on alignment when PPG embeddings are anchors (see Ablation on $\lambda$ in Section \ref{section: extra_ablations}). For the maximum learning rate, we used 1e-3, while batch size was set to 256. As mentioned above, our main encoders are with Transformers whose common hyperparameters are in Appendix Table \ref{table: transformer_hps}, and we perform ablations with EfficientNets whose common hyperparameters are in Appendix Table \ref{table: effnet_hps}.

\begin{figure*}
  \centering \includegraphics[width=1\textwidth]{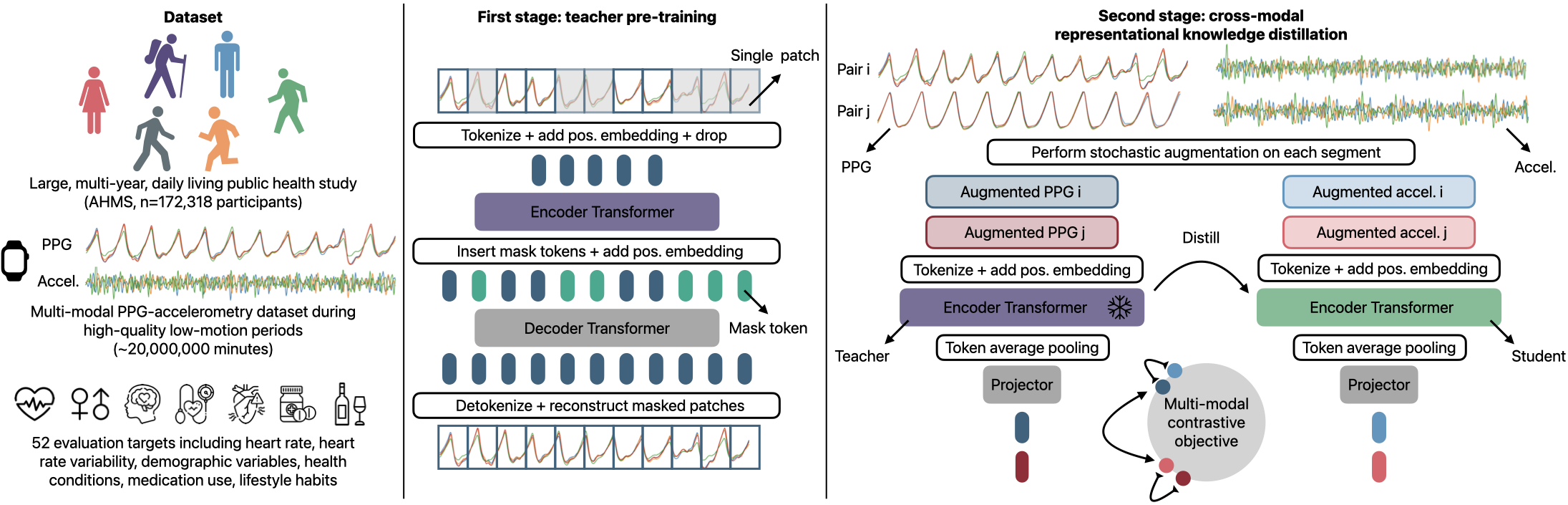}
  \vskip 0.1in
  \caption{Overview of our dataset and methods. We use the multi-modal PPG-accelerometry data collected under informed consent from Apple Watch in Apple Heart and Movement Study. We first pre-train the PPG teacher encoder with masked autoencoding, and then distill its embeddings to an accelerometry encoder via cross-modal knowledge distillation. See Sections \ref{section: teacher_pre_training}, \ref{subsection:methods_kd}, \ref{section: datasets} and \ref{subsubsection: evaluation_metrics} for more details. 
  }
\label{fig: methods}  
\vskip 0.1in
\end{figure*}

\section{Experiments}

\subsection{Datasets}
\label{section: datasets}
We used the PPG and accelerometry signals recorded on Apple Watch from participants in the Apple Heart and Movement Study (AHMS) \citep{macrae_apple_2021, truslow_understanding_2024}. AHMS is an ongoing research study designed to explore the links between physical activity and cardiovascular health, which is sponsored by Apple and conducted in partnership with the American Heart Association and Brigham and Women’s Hospital. To be eligible for the study, participants must be of legal age to provide informed consent (18, 19, or 21 based on location), reside in the United States, have access to an iPhone with the Research app, have an Apple Watch, and provide informed consent within the Research app to participate \citep{shapiro_pulse_2023}.

We created a paired PPG-accelerometry pre-training dataset that we used for pre-training uni-modal and distilled models. Apple Watch intermittently and passively records simultaneous green PPG and accelerometry signals during low-motion periods multiple times per day, to reliably make predictions about an individual's health. PPG and accelerometry signals are recorded simultaneously at a 256Hz or 64Hz sampling rate and are 60 seconds in duration. PPG signals consist of four optical channels corresponding to different combinations of transmitting and receiving diodes, and accelerometry signals consist of 3 channels corresponding to 3 spatial dimensions (see Figure \ref{fig: methods} for signal examples). We curated a pre-training dataset of paired PPG-accelerometry segments from ${\sim}$172K participants, where ${\sim}$20M paired segments were randomly drawn from the full dataset given two conditions: 1) each participant had at least four segments in the pre-training dataset, and 2) the number of segments per participant was as uniform as possible.  PPG segments were pre-processed using dark subtraction (to reject signals introduced by ambient light). Both PPG and accelerometry segments were further pre-processed by bandpass filtering, down-sampling to 64Hz if needed, and temporal channel-wise z-scoring for each segment. For PPG teacher pre-training, we use only the PPG segments of the same dataset. Brief statistics of our curated
dataset are in Table \ref{table: pre_training_dataset}, and AHMS demographics can be found in prior publications \citep{shapiro_pulse_2023, abbaspourazad_large-scale_2024, truslow_understanding_2024}. The training (80\%) and test (20\%) splits were stratified based on participants such that there were no overlapping participants in these two splits. 

\subsection{Evaluation metrics}
\label{subsubsection: evaluation_metrics}

\begin{table}
\small
\centering
\caption{Retrieval analysis for PPG embeddings from accelerometry demonstrates near perfect alignment. Numbers are reported as average (std) across 100 bootstrap candidate pools.}
\vskip 0.1in

\scalebox{0.9}{\begin{tabular}{cccccc}
    \toprule 
    \textbf{Embedding} 
    & Top-1 Acc. $\uparrow$ & Mean Rank $\downarrow$     
 \\
 \midrule
      Accel-KD via PPG-MAE  & \textbf{99.17 (0.23)} & \textbf{1.02 (0.01)} \\
      Accel-MAE + Procrustes align.    & 0.18 (0.03) & 2808.86 (68.95) \\
      Chance-level performance   & 0.01 & 9551.64\\
    \bottomrule
  \end{tabular}}
\label{table: results_retrieval_source_accel}

\vskip 0.1in    
\end{table}

\textbf{PPG-accelerometry pair embeddings retrieval:} To assess the quality of the PPG-distilled accelerometry embeddings (256-D representations after the encoder), we perform a retrieval experiment to see how well the accelerometry embeddings can retrieve their corresponding matched PPG segment on unseen test data. For a batch of paired PPG and accelerometry segments on held-out test participants, we compute the cosine similarity between each distilled accelerometry embedding and PPG embedding, which generates a ranked list that can be used for retrieval. We evaluate retrieval quality using top-1 accuracy for (i.e., for a given query accelerometry embedding, how often is its paired PPG embedding in the most similar embedding). We also report the mean rank, i.e., the mean position of the true paired PPG segment in the rankings; smaller values are better, with 1 indicating perfect retrieval as the correct segment is always ranked first. As an ablation, we repeated the same retrieval analysis from PPG to accelerometry embeddings.

\textbf{Linear probing for downstream targets:} As our main targets, we perform linear probing for predicting heart rate (HR), and two popular measures of heart rate variability: standard deviation of normal-to-normal intervals (SDNN) and root mean square of successive differences (RMSSD). These targets are chosen due to the importance of predicting them frequently thourought the day, and that they are widely used in wearable devices \citep{natarajan_heart_2020} and are indicative of health status \citep{shaffer_overview_2017}, training load in athletes \citep{plews_training_2013} and stress levels \citep{kim_stress_2018}. We also perform linear probing for predicting self-reported age, body mass index (BMI), biological sex, and 46 health targets including health conditions, medication use, and lifestyle habits, for participants in AHMS. More details of our linear probing evaluation metrics is in Appendix \ref{subsection: appendix_dataset}.

\begin{figure*}
  \centering \includegraphics[width=0.85\textwidth]{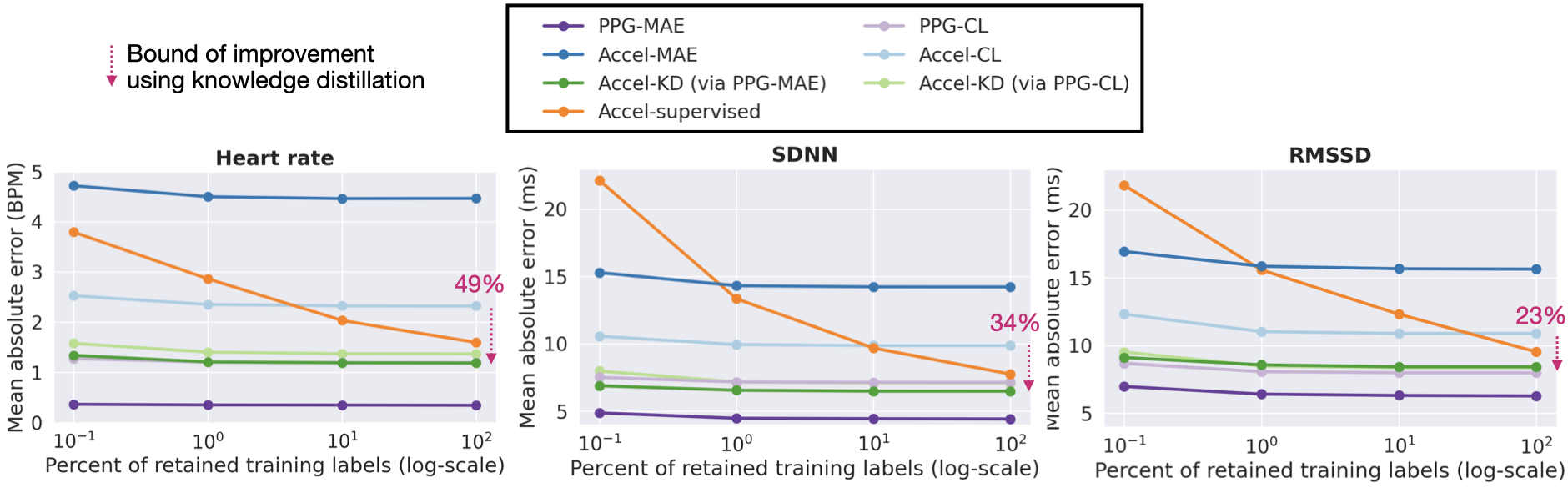}
  \vskip 0.1in
  \caption{Cross-modal representational knowledge distillation improves the quality of accelerometry embeddings. We compare the representational quality of accelerometry encoders via their downstream prediction of heart rate, SDNN and RMSSD. We sweep the number of training segments/labels for supervised training and linear probing training, from 0.1\% to 100\% in the x axis. Distilled accelerometry encoders, ``Accel-KD (via PPG MAE)'' and ``Accel-KD (via PPG-CL)'', are better than their baseline uni-modal accelerometry encoders, ``Accel-MAE'' and ``Accel-CL'', and better than a supervised encoder (``Accel-supervised''). The bound of improvement with the dotted arrows is determined as the difference between the best uni-modal vs. the best distilled accelerometry encoder (Appendix Tables \ref{table: linear_prob_numbers_hr}, \ref{table: linear_prob_numbers_sdnn} and \ref{table: linear_prob_numbers_rmssd}). Compared to the supervised encoder, all pre-trained accelerometry encoders, including distilled ones, demonstrated robustness to the number of available training labels.}
\label{fig: hr_hrv_sweep}  
\vskip 0.1in
\end{figure*}

\section{Results and key takeaways}
\label{section: results}

\subsection{Representational knowledge distillation unlocks strong accelerometry-PPG embedding alignment}
\label{subsection: results_alignment}
We first visually inspected the embeddings of 200 random PPG-accelerometry pairs (20 from 10 participants) from our test split, embedded via PPG teacher encoder trained with MAE (``PPG-MAE'') and 2 accelerometry encoders: 1) accelerometry encoder trained with MAE (``Accel-MAE''), 2) distilled accelerometry encoder (``Accel-KD via PPG-MAE'') from ``PPG-MAE'', by projecting them into 2D t-sne \citep{maaten_visualizing_2008} representation subspace as shown in Appendix Figure \ref{fig: tsne}. We observed a marked difference in the alignment of PPG-accelerometry embeddings for PPG teacher encoder and distilled accelerometry encoder (Appendix Figure \ref{fig: tsne}-right) compared to the uni-modal encoders, visually validating the knowledge transfer from PPG to accelerometry via our representational knowledge distillation framework. 

To quantify the quality of the cross-modality embedding alignment, we performed retrieval analysis from accelerometry embeddings to PPG embeddings (Section \ref{subsubsection: evaluation_metrics}), where PPG embeddings are from the PPG teacher encoder (``PPG-MAE'') and accelerometry embeddings are from the distilled accelerometry encoder (``Accel-KD via PPG-MAE''). To do this, we sample 1,000 random participants from the $\sim$17.5K participants in the test set (no overlap with training), each with an average of 19 PPG and accelerometry segments, and we repeat this procedure 100 times to report standard deviations of our metrics (Table \ref{table: results_retrieval_source_accel}). We observed near-perfect alignment of the PPG and accelerometry embeddings after distillation, indicated by the results in Table \ref{table: results_retrieval_source_accel}: we achieve an average $99.17$ top-1 accuracy and 1.02 mean rank in random batches of $\sim19K$ candidates. We obtain similar strong results when repeating this experiment but switching the retrieval task to be instead from PPG embeddings to accelerometry embeddings (Appendix Table \ref{table: results_retrieval}). To rule out that the near perfect retrieval performance was simply achievable with uni-modal encoder embeddings, we repeated the retrieval analysis using the uni-modal accelerometry embeddings from ``Accel-MAE'' by applying optimal translation, rotation and scaling via Procrustes alignment \citep{krzanowski_principles_2000-1} to make them as close as possible to ``PPG-MAE'' embeddings, and observed marked difference in the retrieval performance (Table \ref{table: results_retrieval_source_accel} and Appendix Table \ref{table: results_retrieval}). Overall, very high retrieval performance (e.g., $99.17$ top-1 accuracy) demonstrates the effectiveness of our representational knowledge distillation framework and how well the distilled accelerometry encoder embeddings match with PPG teacher encoder embeddings.
Importantly, these results indicate that distilled accelerometry embeddings may achieve improved performance for predicting downstream targets due to their high alignment with the high-fidelity PPG embeddings, which we will investigate in the next two sections.

\subsection{Representational knowledge distillation from PPG improves the quality of accelerometry embeddings}
\label{section: results_mae_contrastive_hr}

To gain additional insight about the quality of accelerometry embeddings after knowledge distillation, we compared several encoders in terms of their downstream performance: 1) ``Accel-MAE'', 2) ``Accel-KD via PPG-MAE'', and 3) supervised encoder trained directly on each target from scratch (``Accel-supervised''). In the meantime, to study the label efficiency of these models, we sweep the proportion of available training segments/labels for linear probing and supervised training from 0.1\% to 100\% while keeping the number of test segments/labels the same. Figure \ref{fig: hr_hrv_sweep}a represents the performance of these encoders, as well as the performance of the ``PPG-MAE'' teacher encoder (see Appendix Tables \ref{table: linear_prob_numbers_hr}, \ref{table: linear_prob_numbers_sdnn} and \ref{table: linear_prob_numbers_rmssd} for extended numbers with other metrics). We observed that the distilled accelerometry encoder (``Accel-KD via PPG-MAE'') not only outperformed ``Accel-MAE'', but also was better than the supervised encoder in all targets and all label availability regimes, being particularly robust to the amount of training labels (Figure \ref{fig: hr_hrv_sweep}a across x axis) as it retained strong performance even with $1000\times$ smaller-sized labeled data. This robustness to the amount of labeled data is a major motivation behind keeping our knowledge distillation framework fully unsupervised. 

Additionally, to show that this observation is not unique to the pre-training of the teacher encoder or architecture of the teacher/student encoders, we repeated the pre-training and downstream performance comparison between: 1) accelerometry encoder trained with contrastive learning (``Accel-CL''), 2) distilled accelerometry encoder (``Accel-KD via PPG-CL'') from a PPG teacher encoder trained with CL (``PPG-CL''). To simultaneously show robustness to the architecture, in these experiments, we also changed the teacher and student architectures to EfficientNet in the main results (Section \ref{section: teacher_pre_training}), but conclusions remain the same for contrastive learning with Transformer models (Appendix Table \ref{table: ablation_vit_cl}). We made similar observations with EfficientNet encoders trained via CL: the distilled accelerometry encoder not only outperformed ``Accel-CL'' encoder, but also was better than the supervised encoder in all targets and all label availability regimes (Figure \ref{fig: hr_hrv_sweep}a; see Appendix Tables \ref{table: linear_prob_numbers_hr}, \ref{table: linear_prob_numbers_sdnn} and \ref{table: linear_prob_numbers_rmssd} for extended numbers with other metrics). Interestingly, we made the observation that in uni-modal pre-training, ``PPG-MAE'' was better than ``PPG-CL'', while ``Accel-CL'' was better than ``Accel-MAE'' (Figure \ref{fig: hr_hrv_sweep}a, Appendix Tables \ref{table: linear_prob_numbers_hr}, \ref{table: linear_prob_numbers_sdnn} and \ref{table: linear_prob_numbers_rmssd}), which we further discuss in Appendix \ref{subsection: appendix_discussion}.

\subsection{Distilled accelerometry encoders are predictive of demographic variables and health targets with high accuracy, i.e., they are generalist foundation models}

\begin{table}
\begin{small}
\caption{Distilled accelerometry encoders are predictive of age, BMI and biological sex with high accuracy, and better than baseline accelerometry encoders. Age and BMI metrics are reported with mean absolute error in years and $\text{kg}/\text{m}^2$ respectively, and 
biological sex is reported with ROC AUC. See Appendix \ref{table: conditions_prediction} for other health targets.}
\vskip 0.1in
    \scalebox{0.95}{\begin{tabular}{cccccc}
    \toprule 
    \textbf{Encoder} 
    & Age $\downarrow$ & BMI $\downarrow$ & Biological Sex $\uparrow$    
 \\
 \midrule
      Accel-KD via PPG-MAE  & \textbf{4.04} & \textbf{2.48} & \textbf{0.99} \\
      Accel-KD via PPG-CL    & 4.74 & 2.82 & 0.97 \\
      \midrule
      Accel-MAE    & 7.73 & 3.84 & 0.87 \\
      Accel-CL   & 4.96 & 2.62 & 0.98\\
    \bottomrule
  \end{tabular}}
    \label{table: results_demo}
\end{small}    
\vskip 0.1in
\end{table}

We next questioned whether the distilled accelerometry encoders are predictive of other health-related targets that require capturing waveform information as opposed to pulse timing information that may be sufficient for heart rate and heart rate variability. To this end, we evaluated their downstream prediction performance of a wide array of targets including age, biological sex, BMI and 46 binary health targets derived from AHMS self-reported questionnaires (see Appendix Section \ref{subsection: appendix_results}). We observed that distilled accelerometry encoders are predictive of the demographic variables (Table \ref{table: results_demo}) and health targets (Appendix Table \ref{table: conditions_prediction}), and their prediction performance is better compared to the baseline uni-modal accelerometry encoders (Table \ref{table: results_demo} and Appendix Table \ref{table: conditions_prediction}). This indicates the generalizability of the distilled accelerometry encoders to a wide range of tasks and their capability as a foundation model. To the best of our knowledge, this is the first work demonstrating that a single accelerometry encoder is predictive of demographic variables with high accuracy (Table \ref{table: results_demo}), and is readily predictive of variety of health targets (Appendix Table \ref{table: conditions_prediction}).

\subsection{Cross-modal representational knowledge distillation can also enable model compression}
\label{subsection: model_compression}
One important aspect of models for wearable devices is that they should be compact in size for on-device inference with a minimal power footprint. This allows more frequent estimation of digital biomarkers throughout the day, and makes them available from resource-constrained wearable devices. Therefore, we questioned whether we can perform model compression during the representational knowledge distillation from PPG to accelerometry. To show this, we reduced the size of accelerometry student encoder to 4 new smaller sizes by shrinking the depth and width of the Transformer backbone during the distillation (Appendix Figure \ref{fig: hr_hrv_compression}, and see Table \ref{table: vit_size} for model sizes). We observed that our distillation framework robustly maintains the information quantified by downstream performance, even in encoders with significantly smaller sizes. Particularly, our small encoder (``S'' in Table \ref{table: vit_size}) is still better than the baseline uni-modal accelerometry encoders (``Accel-MAE'' and ``Accel-CL''), while being $\sim5\times$ smaller than the default Transformer model (``XL'' in Table \ref{table: vit_size}).

\subsection{Additional ablation studies}
\label{section: extra_ablations}

Unless otherwise specified, we performed the following ablation studies with ``Accel-KD via PPG-MAE" as it is our main method (Figure \ref{fig: methods}), without loss of generalization given similar qualitative conclusions to ``Accel-KD via PPG-CL" (Section \ref{section: results_mae_contrastive_hr} and Appendix Table \ref{table: ablation_vit_cl}). Additionally, we also performed several other ablations presented in Appendix \ref{subsection: appendix_results}.

\textbf{Augmentations:} During our cross-modal knowledge distillation, both pairs of signals, PPG and accelerometry, are augmented with our stochastic cascade augmentation module (Section \ref{subsection:methods_kd}). While prior works have investigated augmentations for uni-modal contrastive learning of biosignals, the effect of augmentations for multi-modal knowledge distillation of biosignals remains unknown. Therefore, we investigated the importance of augmentations and observed that they were critical for our cross-modal knowledge distillation as shown in Appendix Table \ref{table: augmentation}. For instance, we observed $45\%$ higher mean absolute error for heart rate prediction when the augmentations were absent during the distillation stage. In addition, we investigated the importance of individual augmentation functions during knowledge distillation (Appendix \ref{subsection: appendix_results} and Appendix Table \ref{table: augmentation}).

\textbf{Multi-modal pre-training of both PPG and accelerometry encoders simultaneously results in significantly reduced performance:} As discussed in Section \ref{section: related_work}, there are prior works for multi-modal pre-training of biosignals where they bind different modality embeddings via multi-modal contrastive learning  \citep{deldari_cocoa_2022, thapa_sleepfm_2024}, which does not involve pre-training a uni-modal teacher encoder (stage 1 in Figure \ref{fig: methods}). While, our motivation here is different for we use multi-modal contrastive learning to enable unsupervised representational knowledge distillation, we questioned whether we get the same improvement for the accelerometry encoders when binding PPG and accelerometry embeddings in a multi-modal pre-training setup. Therefore, we did an ablation of multi-modal contrastive learning where both PPG/accelerometry encoders are trainable during training, and to do a fair comparison,  we experimented with different $\lambda$ values. For all different $\lambda$ values, we observed that the learned accelerometry encoder had significantly lower quality embeddings (see Table \ref{table: mm_pretraining} for $\lambda=1$ and Appendix Table \ref{table: mm_pretraining_extra} for other $\lambda$ values) as quantified by a significant drop of performance in all downstream targets; $95\%$, $47\%$ and $35\%$ higher mean absolute error for heart rate, SDNN, and RMSSD, respectively. In addition, we showed that in the multi-modal pre-training, even if we initialize the PPG encoder using the pre-trained weights of ``PPG-MAE'' encoder, we still observe significant degrade in performance (Appendix Table \ref{table: mm_pretraining_extra}). We believe that this is due to the asymmetric amount of information present in the PPG and accelerometry, such that allowing the PPG encoder to update results in more trivial embeddings and degraded performance for the accelerometry encoder. Overall, this observation further demonstrates the importance of two-stage representational knowledge distillation and freezing the teacher encoder to allow maximal knowledge transfer. 

\begin{table}
\centering
\centering
\caption{Ablation on simultaneous PPG-accelerometer multi-modal pre-training compared to two-stage representational knowledge distillation. Numbers are reported in mean absolute error, see Appendix Table \ref{table: mm_pretraining_extra} for additional numbers.}
\vskip 0.1in
\scalebox{0.8}{\begin{tabular}{cccccc}
    \toprule 
    \textbf{Pre-training framework} &
    \textbf{Eval. $\downarrow$} 
    &      
 \\
 \midrule
 \multirow{3}{*}{\begin{tabular}{c@{}p{1.2cm}@{}}Simultaneous multi-modal\\ contrastive learning\end{tabular}}
      & Heart rate (BPM)  &2.36\\
      & SDNN (ms)   &9.68\\
      & RMSSD (ms)   &11.30\\
    \midrule
    
    \multirow{3}{*}{\begin{tabular}{c@{}p{1.2cm}@{}}Cross-modal\\ knowledge distillation\\(ours, Accel-KD via PPG-MAE)\end{tabular}}
      & Heart rate (BPM) &\textbf{1.21}\\
      & SDNN (ms)  &\textbf{6.58}\\
      & RMSSD (ms)   &\textbf{8.40}\\  
    \bottomrule
  \end{tabular}}
\label{table: mm_pretraining}
\vskip 0.1in
\end{table}

\section{Discussion, limitations and future work}
Here, we demonstrated that a single accelerometry encoder can predict heart rate, heart rate variability, demographic variables and a wide array of downstream health targets, i.e., is a foundation model. This was done with a fully unsupervised knowledge distillation framework to transfer knowledge from PPG to accelerometry to improve performance. We showed that our observations are not unique to the teacher/student architecture, or teacher pre-training method, and we can achieve near-perfect alignment of PPG and accelerometry embeddings. In addition, we showed improvements in the quality of the representations compared to self-supervised and supervised baselines, while maintaining label efficiency for downstream targets. We also showed that cross-modal knowledge distillation can enable compressing the student model size. Our work primarily focuses on accelerometry signals during low-motion and sedentary periods (Section \ref{section: datasets}), where accelerometer captures ballistocardiogram and therefore minute cardiovascular-related information useful for health targets \citep{inan_ballistocardiography_2015, kim_ballistocardiogram_2016}. Future work can investigate models that take slower-scale activity metrics on wearable devices such as steps, speed, sleep, and slow changes in accelerometry, as well as minute changes in accelerometry at sedentary settings to improve the performance for downstream targets \citep{hallgrimsson_learning_2018, ni_modeling_2019, spathis_self-supervised_2021,xu_relcon_2024}. In addition, another interesting area of investigation for future work could be experimenting with modality specific augmentations \citep{qian_what_2022, demirel_finding_2023}. One caveat of our work is that it currently supports two modalities, while future work can consider statistical objectives for mutual information maximization, when there are more than one teacher or student modalities \citep{shidani_poly-view_2024}, or by binding all student modalities to a single teacher modality \citep{girdhar_imagebind_2023}. Another caveat is that while this work can be used for knowledge transfer or retrieval of the high-fidelity modality embeddings, it does not provide a generative model across modalities \citep{sarkar_cardiogan_2020}; future work can consider recent techniques to incorporate generative capabilities using unified encoder and decoder Transformers \citep{mizrahi_4m_2023, meta_chameleon_2024}.  Ideally, future work can also consider modeling other modalities such as text, images and videos to leverage information from these other input sources and weakly supervise biosignal representations, similar to prior work for modeling accelerometry during motion and other modalities such as video and text \citep{moon_imu2clip_2022, tan_egodistill_2023}.

\newpage
\section*{Impact Statement}

This work advances machine learning in digital health by enabling the extraction of health insights from accelerometer data commonly available in wearable devices. This could increase the accessibility of health monitoring and improve health outcomes through early detection of health conditions. However, we acknowledge that such models may exacerbate equity gaps between individuals with access to wearable devices and those without. It is also important to address potential ethical concerns, such as privacy and potential biases in model outcomes. We encourage continued research
into interpretability methods suited to foundation models in
healthcare contexts. Responsible development and deployment of these models necessitates careful consideration of these factors to ensure equitable and beneficial outcomes for all individuals. 

\section*{Acknowledgments}
We would like to thank participants in Apple Heart and Movement Study, Calum MacRae, MD, PhD, and study staff at The Brigham and Women's Hospital, a Harvard affiliate, without whom this work would not have been possible. We are grateful for Oussama Elachqar for critical contributions to the data processing, training and evaluation infrastructure. We also thank Sean Jewell and Fahad Kamran for their important contributions to the infrastructure, Nandita Bhaskhar, Jen Block, Chris Brouse, Chris Sandino, and Vimal Thilak for providing valuable feedback regarding the manuscript,  Rachael Cho, Lindsay Hislop, Eduardo Martinez Montes and Laura Rhodes for publication coordination.

\bibliography{library}
\bibliographystyle{icml2025}

\newpage 

\appendix
\onecolumn
\section{Appendix}

\subsection{Implementation details}

The common architectural and training hyperparameters for training Transformer and EfficientNet models, can be found in Tables \ref{table: transformer_hps} and \ref{table: effnet_hps}, respectively. The changes in Transformer model architecture for ablation on compressing the Transformer model size (Section \ref{subsection: model_compression}) can be found in Table \ref{table: vit_size}.

\subsection{Dataset and evaluations}
\label{subsection: appendix_dataset}

\textbf{Dataset statistics}: Brief statistics for our curated PPG-accelerometry dataset from AHMS is available in Table \ref{table: pre_training_dataset}. 

\textbf{Linear probing for heart rate and heart rate variability:} We perform linear probing for predicting heart rate (HR), and two popular measures of heart rate variability: standard deviation of normal-to-normal intervals (SDNN) and root mean square of successive differences (RMSSD). These targets are from Apple Watch's generated values during low-motion periods where PPG peaks are reliably detected, resulting in accurate prediction of heart rate and heart rate variability. These targets are chosen because they are widely used in wearable devices \citep{natarajan_heart_2020} and are indicative of health status \citep{shaffer_overview_2017}, training load in athletes \citep{plews_training_2013} and stress levels \citep{kim_stress_2018}. Being able to predict them via low-fidelity biosignals will enable a more frequent prediction of such targets, giving the users a broader view of their health-related changes in different scenarios of their daily life. We formulate this problem as a regression task, where we use ridge regression to predict the continuous value of these targets and we use mean absolute error to quantify performance. For heart rate, we report error in beats per minute (BPM), and for SDNN and RMSSD, we report error in milliseconds (ms). For these targets given that they change from segment to segment, we perform the linear probing at segment granularity: each segment contributes one and only one sample in the downstream training/evaluation. However, the downstream training/evaluation splits are stratified based on participants such that the evaluation split's participants have no overlap with those in the training split of the linear probing or pre-training. 

\textbf{Linear probing for age, BMI, biological sex, and health targets from AHMS survey questions:} We perform linear probing for predicting self-reported age, body mass index (BMI), biological sex (sex assigned at birth), and health targets from survey questions for participants in AHMS. During AHMS, participants fill out multiple questionnaires containing various questions regarding their historical health record and demographics \citep{truslow_understanding_2024, abbaspourazad_large-scale_2024}. The response to these questions are usually in form of `yes' or `no' for whether the participant has had a history of a health condition (e.g., asthma), or whether they take specific medications (e.g., anti-depressants), or regarding their lifestyle habits (e.g., smoking). For the classification tasks, we use ridge regression to predict scores for binarized targets (0/1) and we quantify the performance with area under curve of receiver's operating curve (AUC). For biological sex, we classify male versus female, and for health targets we classify `yes' versus `no'. For regression tasks (age and BMI), we use ridge regression to predict continuous targets and we use mean absolute error to quantify performance. Age is reported in years and BMI is reported in $\text{kg}/\text{m}^2$. Given that these targets do not vary from segment to segment, we perform the linear probing at participant granularity: we mean-aggregate all the embeddings associated to each participant so that each participant contributes one and only one sample in the downstream training/evaluation. Similar to the heart rate and heart rate variability linear probing, the downstream training/evaluation splits for these targets are stratified based on participants such that the evaluation split's participants have no overlap with those in the training split of the linear probing or pre-training.

AHMS survey is formed of multiple questionnaires which participants fill out over the course of their participation in the study. Tables \ref{table: survey_medical_conditions} and \ref{table: survey_medications} contain AHMS survey questions about medical conditions and medications, respectively, in addition to the corresponding target labels used in Appendix Table \ref{table: conditions_prediction}. Table \ref{table: survey_drinking_smoking} includes AHMS survey questions about drinking and smoking habits, and Table \ref{table: drinking_smoking_logic} defines our logic to summarize these questions into binary labels for the related targets used in Appendix Table \ref{table: conditions_prediction}.

\subsection{Results and ablation studies}
\label{subsection: appendix_results}
\begin{figure}
  \centering \includegraphics[width=1\columnwidth]{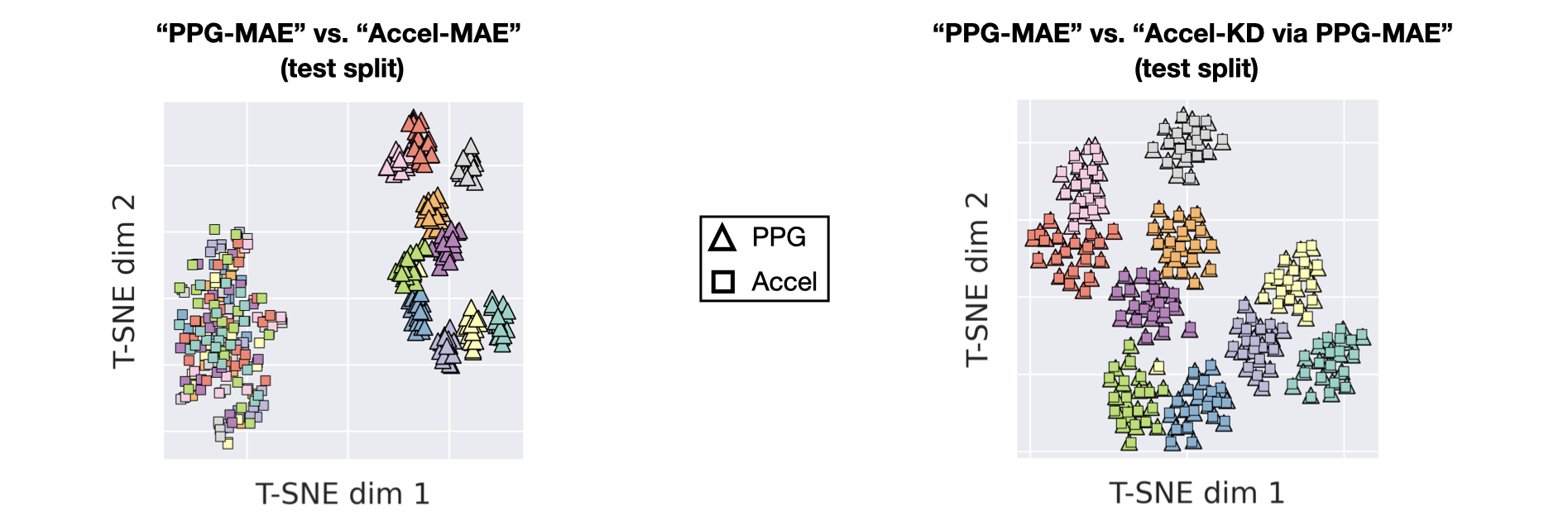}
  \vskip 0.1in
  \caption{2D T-SNE projections of embeddings for PPG pre-trained teacher encoder (``PPG-MAE") and 2 accelerometry encoders: 1) uni-modal encoder pre-trained with masked autoencoding (``Accel-MAE", left), 3) distilled encoder from the PPG teacher encoder (``Accel-MAE via PPG-KD", right). We can visually see marked alignment after distillation in the right panel. Each marker represents an individual segment, where markers are colored based on participants and segments are identical across panels. See retrieval analysis numbers in Table \ref{table: results_retrieval_source_accel}.}
\label{fig: tsne}  
\vskip 0.1in
\end{figure}

\textbf{Visual inspection of the T-SNE representations}: The visual inspection of the T-SNE embeddings for the random, uni-modal and distilled accelerometry encoder as well as the PPG teacher encoder is shown in Figure \ref{fig: tsne} as explained in Section \ref{subsection: results_alignment}.

\begin{figure}
  \centering \includegraphics[width=1\columnwidth]{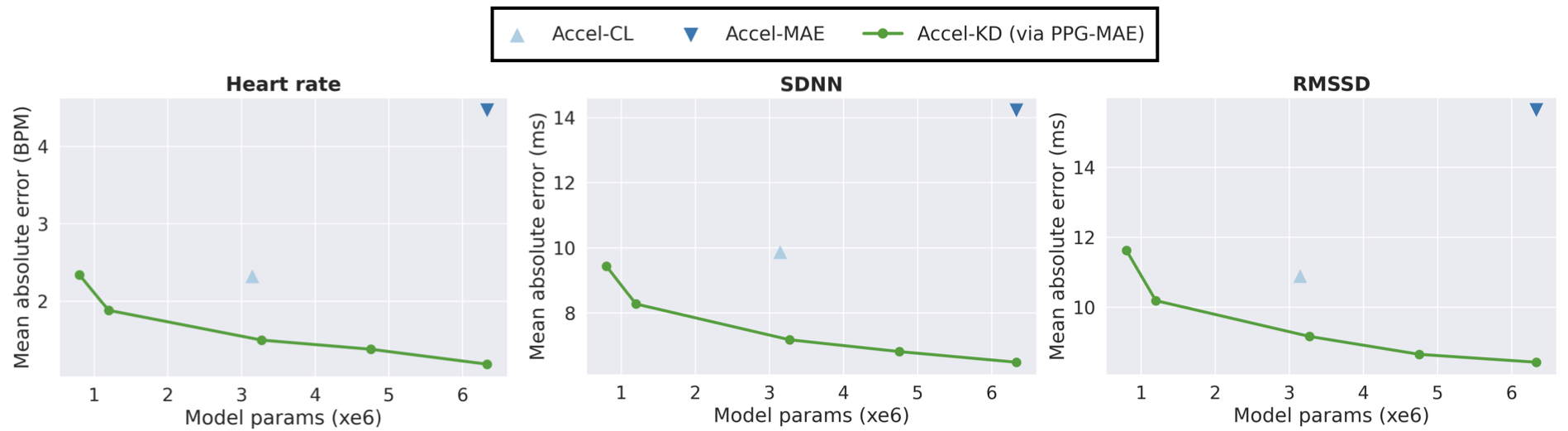}
  \vskip 0.1in
  \caption{Cross-modal representational knowledge distillation can be used for model compression. We show the downstream prediction of heart rate, SDNN and RMSSD while compressing the distilled accelerometry encoder. We observe that even small accelerometry encoders maintain information and are still even better than the baseline accelerometry encoders, while being $\sim5\times$ smaller.}
\label{fig: hr_hrv_compression}  
\vskip 0.1in
\end{figure}

\textbf{Accelerometry model compression via knowledge distillation}: We demonstrate that we can perform model compression for the student accelerometry encoder in Figure \ref{fig: hr_hrv_compression}, as explained in Section \ref{subsection: model_compression}.

\textbf{Retrieval analysis}: Retrieval analysis for accelerometry embeddings from PPG embeddings is shown in Table \ref{table: results_retrieval}.

\textbf{Extended numbers for linear probing of heart rate, SDNN and RMSSD}: Linear probing and supervised evaluation performance numbers at 0.1\% and 100\% data availability in Figure \ref{fig: hr_hrv_sweep}, as well as root mean squared error and Pearson's R metrics, can be found in Appendix Tables \ref{table: linear_prob_numbers_hr}, \ref{table: linear_prob_numbers_sdnn} and \ref{table: linear_prob_numbers_rmssd}.

\textbf{Ablation on choice of positive pairs in teacher pre-training with contrastive learning}: We selected the positive pairs as two augmented views of the same sample. This was done to enforce the encoders to contain more segment-level information necessary for the main downstream targets presented in this study including heart rate and heart rate variability. There are other choices of positive pair selections in prior work, for example participant-level positive pair selection \citep{abbaspourazad_large-scale_2024, pillai_papagei_2024}. Appendix Table \ref{table: ablation_positive_pairs} demonstrates the performance of heart rate, SDNN and RMSSD for the ``PPG-CL" trained with these two different positive pair selection strategies.

\textbf{Ablation on number of PPG channels in teacher pre-training}: We performed an ablation about the effect of the number of PPG channels in teacher pre-training on downstream evaluations in Table \ref{table: ablation_ppg_channels}, where we only kept one of the PPG channels for modeling and observed a drop in performance. This demonstrates the importance of modeling multi-channel PPG signals.

\textbf{Ablation on larger model sizes for teacher pre-training}: We made several optimizations to keep our model sizes small for feasibility on running wearable devices with power and resource constraints. Interestingly, we observed signs of overfitting as we increased our encoder size, which is why our encoder sizes are not larger. This could be due to the fact that one needs to scale the model and data size simultaneously to gain benefits of scaling laws \citep{kaplan_scaling_2020, zhai_scaling_2022}. As an example, when we increased the encoder size in ``PPG-MAE" (from 6.3M to 12.7M) by increasing the number of layers from 8 to 16, we observed initial signs of overfitting as shown in Table \ref{table: ablation_ppg_larger_model}. We believe future work can investigate the scaling laws for encoder models of wearable biosignals by growing the encoder and data size simultaneously \citep{kaplan_scaling_2020, zhai_scaling_2022}.

\textbf{Ablation on augmentations}: In addition to comparing knowledge distillation with and without augmentations, we studied the importance of individual augmentations during knowledge distillation (Section \ref{section: extra_ablations}). To this end, we performed the knowledge distillation from PPG to accelerometry, where we only kept one of the augmentation functions during distillation (applied in every forward pass), one at a time. This was done while maintaining all other training choices the same to control for the effect of augmentations. We observed that: 1) our stochastic augmentation module achieved the highest accuracy (Appendix Table \ref{table: augmentation}), likely because it captures more diverse range of distortions during training, 2) among the individual augmentations, "add Gaussian noise" and "Cut out" had the highest importance, while "Time warp" had the least importance.

In general, our hypothesis for why augmentations are important for knowledge distillation across PPG and accelerometry is that given the relationship between arterial blood flow present in PPG and ballistocardiogram in accelerometry \citep{inan_ballistocardiography_2015, kim_ballistocardiogram_2016}, particularly for aligned PPG-accelerometry segments during low-motion periods which is the focus of our work, knowledge distillation without augmentations is a relatively easier pre-training task compared to that with augmentations. Therefore, we think distillation without augmentations, and even very simple augmentations as shown by individual augmentations results in Appendix Table \ref{table: augmentation}, leads to capturing less minute information, which is relatively similar to why and how the amount and type of augmentations in uni-modal contrastive learning is critical as demonstrated in prior work \citep{chen_simple_2020}.

\textbf{Ablation on impact of $\boldsymbol{\lambda}$ in the cross-modal knowledge distillation} $\mathbf{\lambda}$: We studied the impact of $\lambda$ in Equation \ref{eq: objective} for our representational knowledge distillation. We observed that while the improvements of accelerometry embeddings via distillation were robust to $\lambda$, higher values of $\lambda$ ($\lambda=\{0.75,1\}$) were the most optimal (Appendix Table \ref{table: lambda}), indicating that keeping PPG embeddings as anchor embeddings provided the most knowledge transfer, perhaps due to the fact that PPG is the higher-fidelity modality.

\textbf{Ablation on initialization of the PPG encoder and $\boldsymbol{\lambda}$ in comparison to simultaneous multi-modal pre-training}: In multi-modal pre-training mentioned in Section \ref{section: extra_ablations}, even if we initialize the PPG encoder with PPG-MAE (row 2 of Table \ref{table: mm_pretraining_extra}), the downstream predictions are not as good as knowledge distillation (row 3 of Table \ref{table: mm_pretraining}). We also observed that the conclusions remained in tact with different values of $\lambda$ in the contrastive learning objective in Equation \ref{eq: objective}.

\subsection{Discussion}
\label{subsection: appendix_discussion}
\textbf{Discussion for why in uni-modal pre-training with contrastive learning was better for accelerometry, and with masked autoencoding was better for PPG}: Our hypothesis for this is that given that accelerometry is noisier than PPG (see Figure \ref{fig: methods} for examples), reconstructing accelerometry in the output space via MAE is a rather difficult pre-training method that bottlenecks the quality of extracted representations, as opposed to that for a signal such as PPG which is more structured and less noisier to reconstruct. Therefore, we think MAE pre-training may be more suitable for less noisy biosignals, and CL pre-training is more suitable for noisier biosignals. This, in fact, is a major motivation and difference for pre-training strategies that reconstruct in the representation space versus output space \citep{littwin_how_2024}. All in all, for both pre-training frameworks and Transformer/EfficientNet architectures, our representational knowledge distillation framework robustly distills the information from PPG to accelerometry, and improves the information in the accelerometry embeddings. 

\newpage

\begin{table}
  \centering
  \caption{Common hyperparameters for pre-training experiments involving Transformers.}
  \vskip 0.1in
  \small
\begin{tabular}{lr}
    \toprule 
    \textbf{Hyperparameters} & \textbf{Value}  \\
    \midrule 
    Patch window & 0.3125s (non-overlapping)  \\
    Tokenizer & Learnable linear  \\
    Token dim & 256\\
      Number of layers & 8 \\
      Attention heads & 8 \\
      MLP feedforward hidden dim   &1024 \\
      Normalization   & Layer norm (pre) \\
      Positional embedding   & Sinusoidal \\
      Activation   & GeLU \\
      Output aggregation   & Global average pooling \\
    \midrule
      Optimizer & AdamW\\
      Max learning rate & Variable (refer to text) \\
      Weight decay   &1e-5 \\
      Learning rate scheduling & Warmup for 20K iters, then exp. decay with $\gamma=0.985$ every 1K iters  \\
      Gradient clipping & Max norm at 3  \\
    \bottomrule
  \end{tabular}
  \label{table: transformer_hps}
  \vskip 0.1in
\end{table}

\begin{table}
  \centering
  \caption{Common hyperparameters for pre-training experiments involving EfficientNet \citep{tan_efficientnet_2020}, adapted for time-series as proposed in \citep{abbaspourazad_large-scale_2024}. Here, MBConv1D refers to the 1D version of Mobile Inverted Bottleneck, the standard building block of EfficientNet.}
  \vskip 0.1in
  \small
\begin{tabular}{lr}
    \toprule 
    \textbf{Hyperparameters} & \textbf{Value}  \\
    \midrule 

    \textbf{Encoder Architecture} & \\
    \quad Input layer & Conv1D, BN, Swish \\
    \quad Number of MBConv1D blocks	&  16 \\
    \quad Output layer & Conv1D, BN, Swish, Avg. pooling  \\
    \textbf{MBConv1D} & \\
    \quad Number of Conv1D layers & 5\\
    \quad Expansion factor & 7 \\
    \quad Activation & Swish \\
    \quad Normalization & Batchnorm (BN) \\
    \quad Kernel size &  3 (first 8 blocks), 5 (last 8 blocks) \\
    
    \midrule
      Optimizer & Adam\\
      Max learning rate & 1e-3 \\
      Weight decay   &1e-5 \\
      Learning rate scheduling & Step decay with $\gamma=0.5$ every 125K iters  \\
      Gradient clipping & -  \\
    \bottomrule
  \end{tabular}
  \label{table: effnet_hps}
\vskip 0.1in  
\end{table}

\begin{table}
  \centering
  \caption{Number of participants/segments, average number of calendar days per participant, and total dataset time span (time between the earliest to the latest recorded segment) in our pre-training datasets. We used the data from 80\% of the participants for training and 20\% for test.}
  \small
  \vskip 0.1in
\begin{tabular}{cccc}
    \toprule 
    Number of participants &172,318 \\
      Number of segments   &19,993,427 \\
      Average number of calendar days per participant   &84.12\\
      Total dataset time span (days from 2019-09-21 to 2023-08-29)   &1,439 \\
    \bottomrule
  \end{tabular}
  \label{table: pre_training_dataset}
  \vskip 0.1in
\end{table}

\begin{table}
\centering
\caption{Changes in Transformer architecture parameters when scaling its size. Size names ("XS" to "XL") are picked arbitrarily and are just relative. }
\vskip 0.1in
\small
\begin{tabular}{cccccc}
    \toprule 
    \textbf{Transformer size} &\textbf{XS} 
    & \textbf{S} & \textbf{M}  & \textbf{L} & \textbf{XL}      
 \\
 \midrule
 
Token dim & 128 & 128 & 192 & 256 & 256\\
Number of layers & 4 & 6 & 6 & 6 & 8 \\
      MLP feedforward hidden dim & 512 & 512 & 1024 & 1024 & 1024 \\
      Attention heads & 4 &4 & 6 & 8 & 8 \\
       \midrule
      Number of parameters & 800K & 1.2M & 3.3M & 4.8M & 6.3M \\
    \bottomrule
  \end{tabular}
  \label{table: vit_size}
  \vskip 0.1in
\end{table}

\begin{table}
\centering
\small
\caption{Retrieval analysis for accelerometry embeddings from PPG demonstrates near perfect alignment. Numbers are reported as average (std) across 100 bootstrap candidate pools. We would like to note that due to strong alignment of PPG and accelerometry embeddings in ``Accel-KD via PPG-MAE'', the statistics of retrieval analysis for accelerometry embeddings from PPG embeddings in this table is very similar to that in Table \ref{table: results_retrieval_source_accel} for retrieval analysis of PPG embeddings from accelerometry embeddings.}
\vskip 0.1in
\scalebox{0.95}{\begin{tabular}{cccccc}
    \toprule 
    \textbf{Embedding} 
    & Top-1 Acc. $\uparrow$ & Mean Rank $\downarrow$     
 \\
 \midrule
      Accel-KD via PPG-MAE  & \textbf{99.17 (0.23)} & \textbf{1.02 (0.01)} \\
      Accel-MAE + Procrustes align.    & 0.31 (0.04) & 2311.50 (60.74) \\
      Chance-level performance   & 0.01 & 9551.64\\
    \bottomrule
  \end{tabular}
  }

\label{table: results_retrieval}
\vskip 0.1in
\end{table}

\newpage

\begin{table}
\centering
\caption{Linear probing and supervised evaluation for 100\% (0.1\%) data availability of heart rate (reported in BPM) in Figure \ref{fig: hr_hrv_sweep}a for accelerometry and PPG encoders. Best performance based on 0.1\% data availability is shown in bold, separately for each modality.}
\vskip 0.1in
\small
\begin{tabular}{cccccc}
    \toprule 
     Encoder & Mean absolute error $\downarrow$
    & Root mean squared error $\downarrow$ & Pearson's R  $\uparrow$    
 \\
 \midrule 
Accel-supervised & 1.60 (3.94) & 4.28 (7.76) & 0.87 (0.30) \\
Accel-MAE & 4.47 (4.72) & 7.63 (7.90) & 0.70 (0.67)\\
Accel-CL & 2.33 (2.54) & 4.58 (4.80) & 0.89 (0.88) \\
Accel-KD via PPG-MAE & \textbf{1.19 (1.34)} & \textbf{3.12 (3.20)} & \textbf{0.96 (0.95)} \\
Accel-KD via PPG-CL & 1.37 (1.58) & 3.45 (3.56) & 0.94 (0.93) \\
 \midrule
PPG-MAE & \textbf{0.34 (0.36)} & \textbf{0.55 (0.59)} & \textbf{0.99 (0.98)} \\
PPG-CL & 1.20 (1.28) & 1.63 (1.74) & 0.98 (0.97) \\
    \bottomrule

  \end{tabular}
  \label{table: linear_prob_numbers_hr}
  \vskip 0.1in
\end{table}

\begin{table}
\centering
\caption{Linear probing and supervised evaluation for 100\% (0.1\%) data availability of SDNN (reported in ms) in Figure \ref{fig: hr_hrv_sweep}a for accelerometry and PPG encoders. Best performance based on 0.1\% data availability is shown in bold, separately for each modality.}
\vskip 0.1in
\small
\begin{tabular}{cccccc}
    \toprule 
     Encoder & Mean absolute error $\downarrow$
    & Root mean squared error $\downarrow$ & Pearson's R  $\uparrow$     
 \\
 \midrule 
Accel-supervised & 7.82 (22.38) & 14.86 (32.30) & 0.77 (0.12) \\
Accel-MAE & 14.31 (15.38) & 22.74 (23.80) & 0.47 (0.42) \\
Accel-CL & 9.93 (10.64) & 16.47 (17.23) & 0.72 (0.70) \\
Accel-KD via PPG-MAE & \textbf{6.54  (6.95)} & \textbf{12.62  (13.11)} & \textbf{0.84 (0.83)} \\
Accel-KD via PPG-CL & 7.16 (8.04) & 13.21 (14.04) & 0.82 (0.80) \\
 \midrule
PPG-MAE & \textbf{4.46 (4.92)} & \textbf{7.51 (8.07)} & \textbf{0.94 (0.93)} \\
PPG-CL & 7.17 (7.55) & 11.09 (11.72) &0.87 (0.86) \\
    \bottomrule

  \end{tabular}
  \label{table: linear_prob_numbers_sdnn}
  \vskip 0.1in
\end{table}

\begin{table}
\centering
\caption{Linear probing and supervised evaluation for 100\% (0.1\%) data availability of RMSSD (reported in ms) in Figure \ref{fig: hr_hrv_sweep}a for accelerometry and PPG encoders. Best performance based on 0.1\% data availability is shown in bold, separately for each modality.}
\vskip 0.1in
\small
\begin{tabular}{cccccc}
    \toprule 
     Encoder & Mean absolute error $\downarrow$
    & Root mean squared error $\downarrow$ & Pearson's R  $\uparrow$     
 \\
 \midrule 
Accel-supervised & 9.61 (22.60) & 18.74 (36.01) & 0.70 (0.08) \\
Accel-MAE & 15.70 (17.00) & 27.03 (28.17) & 0.40 (0.35) \\
Accel-CL & 10.95 (12.39) & 19.83 (21.01) & 0.68 (0.64) \\
Accel-KD via PPG-MAE & \textbf{8.47 (9.17)} & \textbf{16.70  (17.33)} & \textbf{0.77 (0.75)} \\
Accel-KD via PPG-CL & 8.42 (9.57) & 16.82 (17.63) & 0.76 (0.74) \\
 \midrule
PPG-MAE & \textbf{6.34 (7.05)} & \textbf{11.18 (12.00)} & \textbf{0.90 (0.88)} \\
PPG-CL & 8.02 (8.71) & 13.67 (14.86) & 0.85 (0.83) \\
    \bottomrule

  \end{tabular}
  \label{table: linear_prob_numbers_rmssd}
  \vskip 0.1in
\end{table}

\begin{table}
\centering
\caption{Downstream target evaluations reported in ROC AUC for AHMS survey questions. We observed that the distilled accelerometry encoders consistently better predicted these targets compared to the uni-modal accelerometry encoders. Asterisks indicate statistical significance for the comparison between the best distilled accelerometry encoder (``Accel-KD via PPG-MAE") versus the best uni-modal accelerometry encoder (``Accel-CL"). For statistical significance, we calculated 200 bootstrapped ROC AUC values for each evaluation, and then performed one-sided Wilcoxon Rank-Sum test to compute the p-value of the statistical comparison. We report ``***" for $P<5\mathrm{e}{-4}$, ``**" for $5\mathrm{e}{-4} \leq P < 5\mathrm{e}{-3}$, ``*" for $5\mathrm{e}{-3} \leq P < 5\mathrm{e}{-2}$ and ``n.s." for $P\geq 5\mathrm{e}{-2}$, where $P$ is the p-value of the comparison.}
\vskip 0.1in
\scalebox{0.85}{\begin{tabular}{lllll}
\toprule
Name & Accel-KD via PPG-MAE & Accel-KD via PPG-CL & Accel-MAE & Accel-CL \\
\midrule
ACE-inhibitors & \textbf{0.802} (***) & 0.794 & 0.731 & 0.791 \\
Active alcohol user & \textbf{0.681} (***) & 0.675 & 0.616 & 0.665 \\
Active smoker & \textbf{0.810} (***) & 0.801 & 0.735 & 0.784 \\
Afib & \textbf{0.816} (***) & 0.804 & 0.765 & 0.798 \\
Allergy & \textbf{0.652} (***) & 0.648 & 0.619 & 0.644 \\
Anti-anxiety & \textbf{0.713} (***) & 0.707 & 0.641 & 0.696 \\
Anti-psychotics & \textbf{0.796} (***) & 0.785 & 0.705 & 0.767 \\
Anticoagulants & \textbf{0.818} (***) & 0.809 & 0.759 & 0.801 \\
Antidepressants & \textbf{0.795} (***) & 0.782 & 0.685 & 0.761 \\
Antiplatelets & \textbf{0.784} (***) & 0.781 & 0.732 & 0.776 \\
Anxiety & \textbf{0.767} (***) & 0.759 & 0.679 & 0.747 \\
Artery disease & \textbf{0.880} (*) & 0.869 & 0.822 & 0.873 \\
Arthritis & \textbf{0.781} (***) & 0.773 & 0.733 & 0.774 \\
Asthma & \textbf{0.634} (***) & 0.630 & 0.596 & 0.621 \\
Beta-blockers & \textbf{0.759} (***) & 0.747 & 0.690 & 0.736 \\
Blood pressure & \textbf{0.798} (***) & 0.789 & 0.732 & 0.787 \\
Blood pressure med. & \textbf{0.710} (***) & 0.697 & 0.651 & 0.694 \\
Calcium-channel blockers & \textbf{0.772} (***) & 0.759 & 0.703 & 0.757 \\
Cancer & \textbf{0.800} (***) & 0.791 & 0.743 & 0.793 \\
Chemotherapy & \textbf{0.735} (***) & 0.704 & 0.626 & 0.714 \\
Cholesterol & \textbf{0.755} (***) & 0.747 & 0.703 & 0.746 \\
Chronic bronchitis & \textbf{0.725} (***) & \textbf{0.725} & 0.683 & 0.714 \\
Depression & \textbf{0.740} (***) & 0.735 & 0.665 & 0.722 \\
Diabetes & \textbf{0.829} (***) & 0.818 & 0.767 & 0.810 \\
Diuretics & \textbf{0.756} (***) & 0.747 & 0.701 & 0.743 \\
Hearing & \textbf{0.719} (***) & 0.713 & 0.676 & 0.709 \\
Heart attack & \textbf{0.835} (*) & 0.832 & 0.771 & 0.831 \\
Heart disease & \textbf{0.857} (***) & 0.845 & 0.801 & 0.843 \\
Heart failure & \textbf{0.857} (n.s.) & 0.838 & 0.789 & 0.855 \\
Heart rhythm & \textbf{0.678} (***) & 0.664 & 0.634 & 0.663 \\
Hip/Knee & \textbf{0.844} (*) & 0.842 & 0.790 & 0.841 \\
Kidney & \textbf{0.694} (***) & 0.687 & 0.646 & 0.678 \\
Liver & \textbf{0.729} (***) & 0.713 & 0.615 & 0.696 \\
Lower back & \textbf{0.685} (***) & 0.681 & 0.651 & 0.674 \\
Neck disorder & \textbf{0.724} (***) & 0.717 & 0.676 & 0.714 \\
Neuropathy & \textbf{0.802} (***) & 0.793 & 0.747 & 0.791 \\
Opioid painkillers & \textbf{0.769} (***) & 0.763 & 0.667 & 0.748 \\
Osteoporosis & \textbf{0.854} (***) & 0.849 & 0.807 & 0.851 \\
Pacemaker & \textbf{0.910} (***) & 0.884 & 0.835 & 0.885 \\
Painkillers & \textbf{0.602} (***) & 0.599 & 0.578 & 0.597 \\
Sleep apnea & \textbf{0.798} (***) & \textbf{0.798} & 0.729 & 0.783 \\
Sleep medication & \textbf{0.673} (***) & 0.661 & 0.622 & 0.652 \\
Stroke or TIA & \textbf{0.790} (n.s.) & 0.779 & 0.743 & 0.789 \\
Thyroid & \textbf{0.750} (***) & 0.746 & 0.712 & 0.743 \\
Urinary & \textbf{0.799} (***) & 0.790 & 0.748 & 0.787 \\
Vision & \textbf{0.657} (***) & 0.655 & 0.627 & 0.651 \\
\bottomrule
\end{tabular}}
\label{table: conditions_prediction}
\vskip 0.1in
\end{table}

\begin{table}
\centering
\caption{Ablation on training contrastive learning based models with Transformer. Here, we report linear probing performance (mean absolute error) for 100\% (0.1\%) data availability equivalent to Figure \ref{fig: hr_hrv_sweep}, and Appendix Tables \ref{table: linear_prob_numbers_hr}, \ref{table: linear_prob_numbers_sdnn} and \ref{table: linear_prob_numbers_rmssd} when the encoder architecture in ``PPG-CL", ``Accel-CL" and ``Accel-KD via PPG-CL" is Transformer. We observed no meaningful difference between these methods when we replaced the architecture and all of our conclusions remains the same.}
\vskip 0.1in
\scalebox{0.8}{\begin{tabular}{cccccc}
    \toprule 
     & Heart rate (BPM) $\downarrow$
    & SDNN (ms) $\downarrow$ & RMSSD (ms)  $\downarrow$     
 \\
 \midrule
Accel-CL (EfficientNet) & 2.33 (2.54) & 9.93 (10.54) & 10.95 (12.39) \\
PPG-CL (EfficientNet) & 1.20 (1.28) & 7.17 (7.55) & 8.02 (8.71) \\
Accel-KD via PPG-CL (EfficientNet) & 1.37 (1.58) & 7.16 (8.04) & 8.42 (9.57) \\
\midrule
Accel-CL (Transformer) & 2.28 (2.48) & 9.81 (10.43) & 10.87 (12.09) \\
PPG-CL (Transformer) & 1.13 (1.20) & 7.23 (7.58) & 8.08 (8.73) \\
Accel-KD via PPG-CL (Transformer)& 1.41 (1.59) & 7.17 (7.92) & 8.67 (9.67) \\
\bottomrule

  \end{tabular}}
\label{table: ablation_vit_cl}
\vskip 0.1in
\end{table}

\begin{table}
\centering
\caption{Ablation on the choice of positive pair selection in teacher pre-training with contrastive learning (``PPG-CL").}
\vskip 0.1in
\scalebox{0.8}{\begin{tabular}{cccc}
    \toprule 
    \textbf{Eval. $\downarrow$} 
    & \textbf{Participant-level positive pairs} & \textbf{Segment-level positive pairs}     
 \\
 \midrule
      Heart rate (BPM) &             2.51 &\textbf{1.21}\\
      SDNN (ms)   &12.24 &\textbf{6.58}\\
      RMSSD (ms)   &11.76 &\textbf{8.40}\\
    \bottomrule
  \end{tabular}}
\label{table: ablation_positive_pairs}
\vskip 0.1in
\end{table}

\begin{table}
\centering
\caption{Ablation on number of PPG channels in teacher pre-training (``PPG-MAE").}
\vskip 0.1in
\scalebox{0.8}{\begin{tabular}{cccc}
    \toprule 
    \textbf{Eval. $\downarrow$} 
    & \textbf{PPG-MAE w/ 1 PPG channels} & \textbf{PPG-MAE w/ 4 PPG channels}          
 \\
 \midrule
      Heart rate (BPM) &             0.39 &\textbf{0.34}\\
      SDNN (ms)   &5.04 &\textbf{4.46}\\
      RMSSD (ms)   &7.30 &\textbf{6.34}\\
    \bottomrule
  \end{tabular}}
\label{table: ablation_ppg_channels}
\vskip 0.1in
\end{table}

\begin{table}
\centering
\caption{Ablation on larger model size for ``PPG-MAE".}
\vskip 0.1in
\scalebox{0.8}{\begin{tabular}{cccc}
    \toprule 
    \textbf{Eval. $\downarrow$} 
    & \textbf{PPG-MAE (12.7M)} & \textbf{PPG-MAE (6.3M)}          
 \\
 \midrule
      Heart rate (BPM) &             \textbf{0.34} &\textbf{0.34}\\
      SDNN (ms)   &5.05 &\textbf{4.46}\\
      RMSSD (ms)   &7.52 &\textbf{6.34}\\
    \bottomrule
  \end{tabular}}
\label{table: ablation_ppg_larger_model}
\vskip 0.1in
\end{table}

\begin{table}
\centering
\caption{Ablation on augmentations during cross-modal knowledge distillation.}
\vskip 0.1in
\scalebox{0.8}{\begin{tabular}{cccccccc}
    \toprule 
    \textbf{Eval. $\downarrow$} 
    & \textbf{W/o augs.} & \textbf{W augs.} & \textbf{Cut out} & \textbf{Gauss. noise} & \textbf{Mag. warp} & \textbf{Channel perm.} & \textbf{T. warp}    
 \\
 \midrule
      Heart rate (BPM) &             1.76 &\textbf{1.21} &1.38 &1.29 &1.76 &1.79 &1.84\\
      SDNN (ms)   &6.98 &\textbf{6.58} &6.96 &6.66 &6.97 &6.77 &10.47\\
      RMSSD (ms)   &8.82 &\textbf{8.40} &8.51 &8.84 &8.96 &8.67 &12.83\\
    \bottomrule
  \end{tabular}}
\label{table: augmentation}
\vskip 0.1in
\end{table}

\begin{table}
\centering
\caption{Ablation on $\lambda$ in the multi-modal contrastive objective (Equation \ref{eq: objective}).}
\vskip 0.1in
\scalebox{0.8}{\begin{tabular}{cccccc}
    \toprule 
    \textbf{Eval. $\downarrow$} 
    & $\boldsymbol{\lambda=0}$ & $\boldsymbol{\lambda=0.25}$ & $\boldsymbol{\lambda=0.5}$ & $\boldsymbol{\lambda=0.75}$ & $\boldsymbol{\lambda=1}$     
 \\
 \midrule
      Heart rate (BPM) & 1.25 &1.26 &1.25 &\textbf{1.19} &1.21\\
      SDNN (ms)  &6.80 &6.69 &6.77 &6.80 &\textbf{6.58}\\
      RMSSD (ms)   &8.78 &8.59 &8.68 &8.78 &\textbf{8.40}\\
    \bottomrule
  \end{tabular}}
\label{table: lambda}
\vskip 0.1in
\end{table}

\begin{table}
\centering
\caption{Ablation on $\lambda$ and not freezing the PPG encoder in simultaneous multi-modal contrastive learning in Table \ref{table: mm_pretraining}.}
\vskip 0.1in
\scalebox{0.8}{\begin{tabular}{cccccc}
    \toprule 
    \textbf{Pre-training framework} &
    \textbf{Eval. $\downarrow$} 
    & $\boldsymbol{\lambda=0}$ & $\boldsymbol{\lambda=0.5}$ & $\boldsymbol{\lambda=1}$     
 \\
 \midrule
 \multirow{3}{*}{\begin{tabular}{c@{}p{1.2cm}@{}}Simultaneous multi-modal contrastive learning \\ (PPG encoder initialized randomly)\end{tabular}}
      & Heart rate (BPM)  &2.81 &2.59 &2.36\\
      & SDNN (ms)   &11.01 &10.37 &9.68\\
      & RMSSD (ms)   &12.55 &11.90 &11.30\\
    \midrule
    \multirow{3}{*}{\begin{tabular}{c@{}p{1.2cm}@{}}Simultaneous multi-modal contrastive learning \\ (PPG encoder initialized with ``PPG-MAE" weights)\end{tabular}}
      & Heart rate (BPM) &2.73 &2.61&2.23\\
      & SDNN (ms)  &10.97 &10.29 &9.63\\
      & RMSSD (ms)   &12.56 &11.89 &11.06\\
    \midrule  
    \multirow{3}{*}{\begin{tabular}{c@{}p{1.2cm}@{}}Cross-modal knowledge distillation\\(ours, Accel-KD via PPG-MAE)\end{tabular}}
      & Heart rate (BPM) &1.25 &1.25&\textbf{1.21}\\
      & SDNN (ms)  &6.80 &6.77 &\textbf{6.58}\\
      & RMSSD (ms)   &8.78 &8.68 &\textbf{8.40}\\  
    \bottomrule
  \end{tabular}
  }
\label{table: mm_pretraining_extra}
\vskip 0.1in
\end{table}

\begin{table}
\centering
\caption{AHMS survey questions about medical conditions. The main question is in form of 'Have you ever been diagnosed with any of the following conditions?' and participants can answer 'Yes' or 'No' or 'I prefer not to answer' or 'I don't know'. The question for vision and hearing loss is different, which we explicitly mention in the corresponding rows. Third column indicates the number of left out participants for evaluation -- the reason for variations is that for each target we exclude participants whose answers were 'I prefer not to answer' or 'I don't know' or missing.}
\vskip 0.1in
\scalebox{0.8}{\begin{tabular}{|p{3cm}|p{8.4cm}|p{1.2cm}|} 
    \hline
    \textbf{Target label} & \textbf{Medical condition} & \textbf{N (test)}\\
    \hline
    Heart attack & Heart attack (myocardial infarction) & 26,806\\
    \hline
    Heart disease & Coronary heart disease or angina pectoris & 26,584\\
    \hline
    Blood pressure & High blood pressure (hypertension)& 26,326\\
    \hline
    Stroke or TIA  & Stroke (cerebral hemorrhage, cerebral thrombosis) or transient ischemic attack (ministroke) & 26,805\\
    \hline
    Afib & Atrial fibrillation & 26,342\\
    \hline
    Heart rhythm & Heart rhythm problem other than atrial fibrillation & 26,068\\
    \hline
    Pacemaker & Pacemaker & 26,916\\
    \hline
    Artery disease & Peripheral artery disease & 26,429\\
    \hline
    Heart failure & Heart failure & 26,843\\
    \hline
     Diabetes & Diabetes & 26,661\\
    \hline
    Cholesterol & High cholesterol & 26,238\\
    \hline
    Arthritis & Arthritis & 26,429\\
    \hline
    Hip/Knee & Hip or knee replacement & 26,948\\
    \hline
    Lower back & Low back disorder or other chronic back defect & 26,480\\
    \hline
    Neck disorder & Neck disorder or other chronic neck defect & 26,628\\
    \hline
    Osteoporosis & Osteoporosis & 26,554\\
    \hline
    Asthma & Asthma & 26,762\\
    \hline
    Chronic bronchitis & Chronic bronchitis, chronic obstructive pulmonary disease, or emphysema & 26,707\\
    \hline
    Allergy & Rhinitis, hay fever, eye inflammation, dermatitis, food allergy or other allergy (allergic asthma excluded) & 26,626\\
    \hline
    Kidney & Kidney problems & 26,640\\
    \hline
    Thyroid & Thyroid disease & 26,502\\
    \hline
    Cancer & Cancer & 26,783\\
    \hline
    Liver & Cirrhosis of the liver & 26,784\\
    \hline
    Urinary & Urinary incontinence & 26,728\\
    \hline
    Neuropathy & Neuropathy & 26,370\\
    \hline
    Depression & Depression & 26,110\\
    \hline
    Anxiety & Anxiety disorder & 25,989\\
    \hline
    Hearing & Do you have hearing loss? & 25,633\\
    \hline
    Vision & Do you have vision loss? & 25,895\\
    \hline
  \end{tabular}}
\label{table: survey_medical_conditions}
\vskip 0.1in
\end{table}

\begin{table}
\centering
\caption{AHMS survey questions about medications. The main question is in form of 'Do you currently take any of the following types of medications?' and participants can answer 'Yes' or 'No' or 'I prefer not to answer'. The formatting for the medications is similar to their presentation in the study, but may not exactly match the format in the study application. Third column indicates the number of left out participants for evaluation -- the reason for variations is that for each target we exclude participants whose answers were 'I prefer not to answer' or missing. Third party trademarks used herein are trademarks of their respective owners.}
\vskip 0.1in
\scalebox{0.8}{\begin{tabular}{|p{3cm}|p{8.4cm}|p{1.2cm}|} 
    \hline
    \textbf{Target label} & \textbf{Medications} & \textbf{N (test)} \\
    \hline
    ACE-inhibitors & ACE-inhibitors or ARBs (for blood pressure) such as captopril, enalapril, lisinopril, losartan, ramipril, or valsartan & 11,043\\
    \hline
    Anti-anxiety & Anti-anxiety aids such as alprazolam (Xanax\textsuperscript{\textregistered}), clonazepam (Klonopin\textsuperscript{\textregistered}), clorazepate (Tranxene\textsuperscript{\textregistered}), diazepam (Valium\textsuperscript{\textregistered}), or lorazepam (Ativan\textsuperscript{\textregistered}) & 15,876\\
    \hline
    Anti-psychotics & Anti-psychotics such as haloperidol (Haldol\textsuperscript{\textregistered}), aripiprazole (Abilify\textsuperscript{\textregistered}), risperidone (Risperdal\textsuperscript{\textregistered}), quetiapine (Seroquel\textsuperscript{\textregistered}), olanzapine (Zyprexa\textsuperscript{\textregistered}), clozapine (Clozaril\textsuperscript{\textregistered}), or lurasidone (Latuda\textsuperscript{\textregistered}) & 15,914\\
    \hline
    Anticoagulants  & Anticoagulants (blood thinners) such as warfarin (Coumadin\textsuperscript{\textregistered}), apixaban (Eliquis\textsuperscript{\textregistered}), betrixaban (Bevyxxa\textsuperscript{\textregistered}), dabigatran (Pradaxa\textsuperscript{\textregistered}), edoxaban (Lixiana\textsuperscript{\textregistered}), or rivaroxaban (Xarelto\textsuperscript{\textregistered}) & 15,906\\
    \hline
    Antidepressants & Antidepressants such as amitriptyline (Elavil\textsuperscript{\textregistered}), bupropion (Wellbutrin\textsuperscript{\textregistered}), citalopram (Celexa\textsuperscript{\textregistered}), duloxetine (Cymbalta\textsuperscript{\textregistered}), escitalopram (Lexapro\textsuperscript{\textregistered}), fluoxetine (Prozac\textsuperscript{\textregistered}), paroxetine (Paxil\textsuperscript{\textregistered}), mirtazapine (Remeron\textsuperscript{\textregistered}), sertraline (Zoloft\textsuperscript{\textregistered}), or venlafaxine (Effexor\textsuperscript{\textregistered}) & 15,919\\
    \hline
    Antiplatelets & Antiplatelets (blood thinners) such as aspirin, clopidogrel (Plavix\textsuperscript{\textregistered}), prasugrel (Effient\textsuperscript{\textregistered}), or ticagrelor (Brilinta\textsuperscript{\textregistered}) & 15,891\\
    \hline
    Beta-blockers & Beta-blockers (for blood pressure or heart rhythm) such as atenolol (Tenormin\textsuperscript{\textregistered}), bisoprolol (Zebeta\textsuperscript{\textregistered}), carvedilol (Coreg\textsuperscript{\textregistered}), labetalol, metoprolol (Lopressor\textsuperscript{\textregistered}, Toprol-XL\textsuperscript{\textregistered}), nadolol (Corgard\textsuperscript{\textregistered}), nebivolol (Bystolic\textsuperscript{\textregistered}), propranolol (Inderal\textsuperscript{\textregistered}), or sotalol (Betapace\textsuperscript{\textregistered}) & 15,868\\
    \hline
    Blood pressure med. & Other medications for lowering blood pressure such as clonidine, hydralazine, minoxidil, or sacubitril/valsartan (Entresto\textsuperscript{\textregistered}) & 15,835\\
    \hline
    Calcium-channel blockers & Calcium-channel blockers (for blood pressure or heart rhythm) such as amlodipine (Norvasc\textsuperscript{\textregistered}), diltiazem, or verapamil & 15,812\\
    \hline
     Chemotherapy & Certain types of chemotherapy such as carboplatin, cisplatin, oxaliplatin, vincristine, or vinblastine & 11,084\\
    \hline
    Diuretics & Diuretics (water pills) such as chlorthalidone, furosemide (Lasix\textsuperscript{\textregistered}), hydrochlorothiazide, or spironolactone & 15,929\\
    \hline
    Opioid painkillers & Opioid painkillers such as codeine, fentanyl, hydrocodone, hydromorphone (Dilaudid\textsuperscript{\textregistered}), meperidine (Demerol\textsuperscript{\textregistered}), morphine, oxycodone, Percocet\textsuperscript{\textregistered}, or Vicodin\textsuperscript{\textregistered} & 15,941\\
    \hline
    Painkillers & Non-steroidal anti-inflammatories (painkillers) such as aspirin, celecoxib (Celebrex\textsuperscript{\textregistered}), diclofenac (Cambia\textsuperscript{\textregistered}), ibuprofen (Motrin\textsuperscript{\textregistered}/Advil\textsuperscript{\textregistered}), or naproxen (Aleve\textsuperscript{\textregistered}) & 15,919\\
    \hline
    Sleep medication. & Sleeping aids such as eszopiclone (Lunesta\textsuperscript{\textregistered}), zaleplon (Sonata\textsuperscript{\textregistered}), or zolpidem (Ambien\textsuperscript{\textregistered}) & 15,892\\
    \hline
  \end{tabular}}
\label{table: survey_medications}
\vskip 0.1in
\end{table}

\begin{table}
\centering
\caption{AHMS survey questions about drinking and smoking habits. These are standardized questions from the AUDIT-C questionnaire \citep{bush_audit_1998} and All of US research program \citep{denny_all_2019, ramirez_all_2022}.}
\vskip 0.1in
\scalebox{0.8}{\begin{tabular}{|p{1cm}|p{5cm}|p{5.5cm}|} 
    \hline
    \textbf{\#} & \textbf{Question} & \textbf{Answer choices} \\
    \hline
    Q1 & In your entire life, have you had at least 1 drink of any kind of alcohol, not counting small tastes or sips? & 'Yes'/'No'/'I don’t know'/'I prefer not to answer' \\
    \hline
    Q1b & [If yes to Q1] How often did you have a drink containing alcohol in the past year? & 'Never'/'Monthly or less'/'Two to four time a month'/'Two to three times a week'/'Four or more times a week'/'I prefer not to answer'\\
    \hline
    Q1c & [If yes to Q1] On a typical day when you drink, how many drinks do you have? & '1 or 2'/'3 or 4'/'5 or 6'/'7 to 9'/'10 or more'/'I prefer not to answer'\\
    \hline
    Q2 & Have you smoked at least 100 cigarettes in your entire life? & 'Yes'/'No'/'I don’t know'/'I prefer not to answer'\\
    \hline
    Q2b & [If Yes or Do not know to Q2] Do you now smoke cigarettes every day, some days, or not at all? & 'Every day'/'Some days'/'Not at all'/'I prefer not to answer'\\
    \hline
  \end{tabular}}
\label{table: survey_drinking_smoking}
\vskip 0.1in
\end{table}

\begin{table}
\centering
\caption{Our logic for defining drinking and smoking targets from questions in Table \ref{table: survey_drinking_smoking}. Third column indicates the number of left out participants for evaluation -- the reason for variations is that for each target we exclude participants whose answers did not conclude in our binary 'yes' or 'no' mappings or were missing.}
\vskip 0.1in
\scalebox{0.8}{\begin{tabular}{|p{3cm}|p{8.4cm}|p{1.2cm}|} 
    \hline
    \textbf{Target label} & \textbf{Logic} & \textbf{N (test)}\\
    \hline
    Active alcohol user & 'Yes': answer to Q1 is 'Yes' and answer to Q1b is 'Two to three times a week'/'Four or more times a week' & 23,472\\ &
    'No': answer to Q1 is 'No', or answer to Q1b is 'Never'/'Monthly or less'/'Two to four time a month' & \\
    
    \hline
    Active smoker & 'Yes': answer to Q2 is 'Yes' and answer to Q2b is 'Every day'/'Some days' & 23,452\\ &
    'No': answer to Q2 is 'No', and answer to Q2b is 'Not at all' & \\
    \hline
  \end{tabular}}
\label{table: drinking_smoking_logic}
\vskip 0.1in
\end{table}

\end{document}